# Arabic Text Sentiment Analysis: Reinforcing Human-Performed Surveys with Wider Topic Analysis

**Latifah Almuqren[1,2], Ryan Hodgson[2] and Alexandra I. Cristea[2]**

[1] Department of Information System, College of Computer and Information Sciences, Princess Nourah bint Abdulrahman University, P.O. Box 84428, Riyadh 11671, Saudi Arabia.

[2] Computer Science department, University of Durham, Durham, United Kingdom

Corresponding author: Alexandra I. Cristea (e-mail: alexandra.i.cristea@durham.ac.uk).

**ABSTRACT** Sentiment analysis (SA) has been, and is still, a thriving research area. However, the task of Arabic sentiment analysis (ASA) is still underrepresented in the body of research. This study offers the *first in-depth and in-breadth analysis of existing ASA studies of textual content* and identifies their common themes, domains of application, methods, approaches, technologies and algorithms used. The in-depth study *manually* analyses 133 ASA papers published in the English language between 2002 and 2020 from four academic databases (SAGE, IEEE, Springer, WILEY) and from Google Scholar. The in-breadth study uses modern, automatic *machine learning techniques*, such as topic modelling and temporal analysis, on Open Access resources, to reinforce themes and trends identified by the prior study, on 2297 ASA publications between 2010-2020. The main findings show the different approaches used for ASA: *machine learning*, *lexicon-based* and *hybrid approaches*. Other findings include ASA 'winning' algorithms (*SVM*, *NB*, *hybrid methods*). *Deep learning* methods, such as LSTM can provide higher accuracy, but for *ASA sometimes the corpora are not large enough to support them*. Additionally, whilst there are some ASA corpora and lexicons, more are required. Specifically, Arabic tweets corpora and datasets are currently only moderately sized. Moreover, Arabic lexicons that have high coverage contain only *Modern Standard Arabic* (MSA) words, and those with Arabic dialects are quite small. Thus, *new corpora need to be created*. On the other hand, ASA tools are stringently lacking. There is a *need to develop ASA tools that can be used in industry, as well as in academia, for Arabic text SA*. Hence, our study offers insights into the challenges associated with ASA research and provides suggestions for ways to move the field forward such as lack of Dialectical Arabic resource, Arabic tweets, corpora and data sets for sentiment analysis.

## I. INTRODUCTION

With the advent of big data and the proliferation of social media (e.g., Facebook, Twitter), Sentiment Analysis (SA) has observed an increase in academic research over the past decade. SA, or 'opinion mining', refers to the computational processing of opinions, feelings and attitudes towards a particular event or issue [1, 2]. To identify subjective opinions, SA applies natural language processing (NLP) and textual analytics techniques [3, 4]. SA helps to reveal the polarity of texts, through identifying whether a fragment of text indicates, e.g., a positive, negative or neutral impression [5, 6]. This work also represents the grounding for a series of work on predicting the satisfaction of Telecom companies' customers [7-9]. Next, we consider SA application areas, types, before dwelling into the specific challenges of Arabic SA, the very lightweight related research leading to our research questions.

### A. SENTIMENT ANALYSIS APPLICATION AREAS

Social media sites have become popular in recent years, and since then, the SA approach has grown to become prominent for capturing public opinion. Doing so can improve the effectiveness and efficiency of decision making, through the use of textual analytics, to make better-informed decisions [2, 10]. SA has been adopted in a wide range of fields, including marketing and e-commerce, customer relationship management, market intelligence, strategic planning, political polls, employment, sociology, healthcare, education and scientific research, and humanitarian assistance and disaster relief [2]. SA plays a vital role in obtaining realistic information related to public opinion. For example, SA assisted in determining the preferences of customers and evaluated their satisfaction with products on e-commerce sites like Amazon, with this contributing to improving quality and standards based on the actual needs of customers [11].

SA has been applied from different perspectives, either for general [3, 12, 13] or specific challenges [2, 14, 15] techniques [4, 16] and languages [15, 17, 18].

### B. SENTIMENT ANALYSIS TYPES

Subjectivity Analysis can classify a text as *subjective* or *objective*, where subjective text contains opinions and sentiment, and objective text has facts [19]. SA can classify the text in many ways, i.e., *binary classification* (positive or negative), or *three-way classification* (positive, neutral or negative) [20, 21]. Moreover, SA can be applied via two main approaches, *flat classification*, or *hierarchal classification*. In a flat classification, the classifier classifies the text in a many-way classification on one level; whereas a hierarchical classification uses many layers; usually, in the first layer, the text is classified into (objective, subjective), then the subjective text is classified according to its polarity.

SA has been investigated at three different levels: *document level*, *sentence level*, and *entity or aspect level* [22]. Document level considers the whole text as one unit holding one opinion, such as for product reviews. While sentence level deals with each sentence as a one unit, holding sentiment. Usually, sentence level classification is applied to short texts in social media, such as Twitter [23, 24]. For the aspect level, SA is performed towards an entity aspect [25].

### C. SPECIFIC ARABIC SA CHALLENGES

In the context of language, the majority of the research pertains to English rather than Arabic SA (ASA) [26, 27]. Both languages differ in their expressive power of sentiments, which makes the detection of sentiment polarity considerably more complex [12]. This issue is particularly challenging, given that natural languages are unstructured, rendering interpretation of sentiment a tiresome task [28]. Important studies have handled this problem in the English language [2, 13, 29], but it remains largely unexplored with regards to Arabic [12], even if the latter is ranked as the fourth language used on the Internet, with 270 million users in 2017.

Arabic differs from English in several key aspects. Arabic is a rich morphological language [30, 31], written from right to left, using different forms, thus presenting researchers with specific challenges. Arabic language has many forms, which are Classical Arabic (CA), as in the book of Islam's Holy Quran, Modern Standard Arabic (MSA) used in newspapers, education and formal speaking, Dialectical Arabic (DA) which is the informal everyday spoken language, found in chat rooms and social media platforms. The Arabic language consists of 28 Arabic alphabet letters. Additionally, there are 10 letters with a second form [32], Table 1. To represent the meaning of an Arabic word, diacritics are used, which are small signs over or under letters, positioned to reflect the vocals. Absence of these Arabic diacritics in the DA text makes the text interpretation more complicate. Moreover, DA forms differ from one Arab country to another, rendering understanding specific DA difficult for the people not speaking that DA. [33] have defined six Arabic dialects, which are Gulf, Yemeni, Iraqi, Egyptian, Levantine and Maghrebi.

TABLE 1
ARABIC LETTERS [60]

| | |
|---|---|
| 28 Arabic alphabet letters | أ,ب,ت,ث,ج,ح,خ,د,ذ,ر,ز,س,ش,ص,ض,ط, ظ,ع,غ,ف,ق,ك,ل,م,ن,ه,و,ي |
| 10 letters with a second form | ء(أ,إ,ء,ى,ئ,ؤ) |
| | هـ(ه,ة) |
| | ا(ى,آ) |
| Basic diacritics | ̇ ,̈ , |

This makes the ASA process more complex, for instance, when attempting to build an Arabic lexicon [30, 34-36]. In addition, the mix of MSA and dialectical Arabic employed by

Internet users [37, 38] presents challenges, which have resulted in limited research on ASA [39, 40].

Although ASA is of growing importance, it is still in the early stages of research [27, 41, 42]. Early ASA research addressed SA in newswires [43, 44], whereas the most recent studies focus more on ASA for social media [42, 45-48].

### D. RELATED RESEARCH: ARABIC SA SURVEYS

Furthermore, similarly to the SA implementations, although many survey studies extensively address SA in the English language [29, 49][26, 46-49], ASA survey research is still relatively modest [41]. Some ASA research addresses specific issues, such as creating an Arabic lexicon [50, 51], while others focus only on specific SA techniques [6, 52-54]. A recent study focussed specifically on the popular Deep Learning approaches in ASA, for subjective sentiment analysis [55]. These studies however provide narrow insights into ASA; they do not comprehensively address ASA in general [41]. Thus, the ASA state-of-the-art and progress remains largely unexplored [47], which we are addressing with this study.

### E. RESEARCH QUESTIONS AND CONTRIBUTIONS

Therefore, this paper offers an in-depth analysis of existing ASA studies of textual content and identifies their *common themes*, *domains of application*, *methods*, *approaches*, *technologies*, and *algorithms* used. ASA papers published in the English language between 2002 and 2020 were collected from four academic databases and from Google Scholar. Papers were screened and analysed, with results of screening identifying 133 papers related directly to Arabic text SA. Their contents were manually analysed, and our study presents the different approaches used to conduct this analysis. Additionally, an in-breadth analysis is further performed automatically over the period 2010-2020, involving 2297 ASA publications identified through available API resources. This automatic, large-scale analysis contributes with a quantitative support of areas identified through our manual analysis, such as the *presence of dialect types within ASA* or *popularity of corpus sources*.

Hence, we have identified the following research questions, as a first step towards our systematic ASA review:

*RQ1. What is the current state of research related to ASA?*
*RQ2. What are the most effective approaches, tools and resources used in ASA?*
*RQ3. Which are the most significant challenges in the reviewed ASA studies?*
*RQ4. What are suggestions to overcome the ASA challenges?*

Thus, the main contributions of this work are:
• To the best of our knowledge this is the *first study of this type, which provides a long-term in-depth study of ASA, as well as a large-scale in-breadth analysis of ASA*. We thus provide the first comprehensive depth-and-breadth review of ASA studies published in the current literature.

• Secondly, we present a holistic view of the most significant approaches, tools and resources used in ASA research. Transfer learning is still new to the ASA filed, with only one study found in 2019 and one study found in 2020.

• Thirdly, we assess the most significant challenges in the reviewed studies and propose suggestions to overcome the challenges. Arabic lexicons with high coverage contain only MSA words, and those with Arabic dialects are quite small. New corpora need to be created.

The next section presents the methods applied to identify relevant studies. Section 3 presents their synthesis and discussion. Section 4 presents the conclusions of our review.

## II. METHODS

### A. LITERATURE SEARCHES

In order to identify relevant studies, we systematically reviewed the literature (as recommended by [56]) across four academic databases (SAGE, IEEE, Springer, WILEY) and on Google Scholar, up to June 2020. We used the following keywords within our database search procedures: 'Arabic semantic analysis', 'Arabic subjective analysis', 'Arabic emotion detection', 'Arabic text categorization', 'Arabic opinion mining', 'Arabic lexicon, Arabic corpora', 'Arabic sentiment analysis', 'Arabic sentiment classification', and 'Arabic Opinion Mining'. Some terms were excluded, such as 'Arabic indexing', 'information retrieval' and 'code-switching', to ensure collecting the papers on the appropriate topics, as these were not referring to sentiment analysis.

### B. STUDY SELECTION

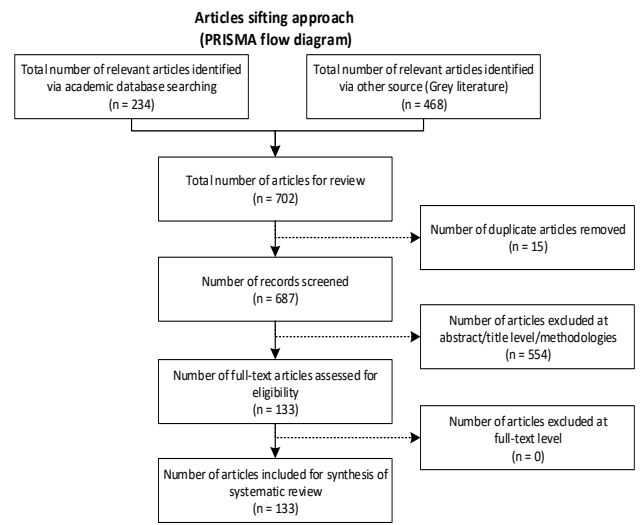

One reviewer (LT) applied the following manual, careful search and review strategy:
1. LT reviewed the studies manually at the title and abstract level, after eliminating the duplicates.
2. The remaining articles were evaluated in detail at full-text level and were included if the reviewer identified them as relevant. The appraisal was carried out using the following inclusion criteria: selected studies:
   (a) reported the application of text ASA, and
   (b) were written in the English language. Studies were excluded if they focused on topics other than text ASA - such as speech, voice or images.

### C. DATA EXTRACTION AND ANALYSIS
Data extraction was performed, during the review of the included studies and ASA methods, as further detailed in Section III.A and Figure 1. The extracted data are synthesised and presented in Section III.B. The information gathered from these syntheses was used to find the common themes in this review.

### C. TOPIC MODELLING METHODOLOGY
When applying Latent Dirichlet Allocation (LDA)[57] to the automatically identified literature, we perform optimisation of parameters through evaluation of coherence score [58] at varying numbers of topics. As LDA requires specification of the total number of topics, we performed an exhaustive search to identify the ideal number of topics, aiming to maximise coherence. The topic word-distributions for the best performing models at different levels of alpha were then collected and analysed at a qualitative level. The optimum model and associated parameters are presented in Appendix V. We evaluate resulting topics and then build a topic-word list which we apply to searching temporal topic frequency within our automatically obtained corpus.

## III. RESULTS

### A. SEARCHES AND SIFTING
702 potential candidate studies were identified via the search strategies, with 687 studies remaining following removal of duplicates. Step 1 of the search and review strategy included the screening of titles, abstracts and methodologies of the 687 studies. A total of 554 studies were thus removed, as these did not meet the inclusion criteria. Step 2 of the review consisted of a detailed assessment of the remaining 133 studies. No studies were excluded at this stage, as all met the inclusion criteria set for this review. The result from the two stages of sifting is presented in a PRISMA diagram (the reporting components most used for systematic reviews) in Figure 1. In order to enhance readability and reduce diagram complexity, guidelines [59] were referred to when designing the diagram.

**FIGURE 1** PRISMA diagram of the ASA Literature Filtering Process [244]

Our first stage (Step 1) of screening revealed that the most common terms related to the topics were *Arabic*, *sentiment*, *mining*, and *opinion*. Conversely, our second stage (Step 2) focused heavily on languages. This analysis further indicates the domains involving ASA research, i.e., *finance* and *news*, and the topics of concern in these papers, which were *linguistics*, *corpora* and *lexicons* (Figure 2a, 2b) and other topics not further discussed in this survey, such as *feature engineering*, with the latter being less prominent in the Arabic language processing, as opposed to other languages.

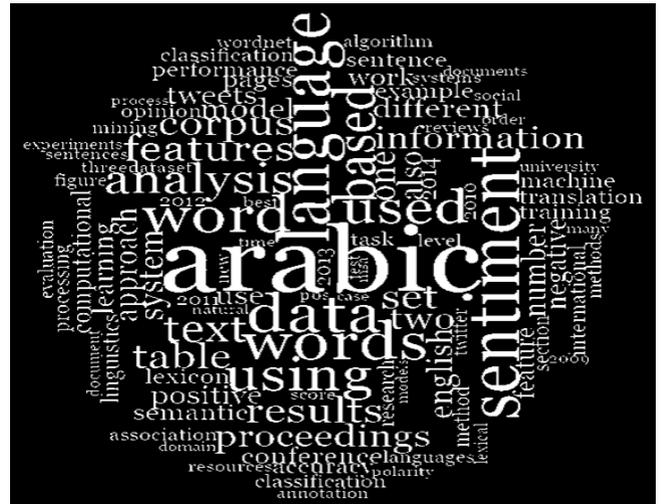

**FIGURE 2a** Word cloud depicting the most frequent words appearing in Step 1 of the ASA screening [244].

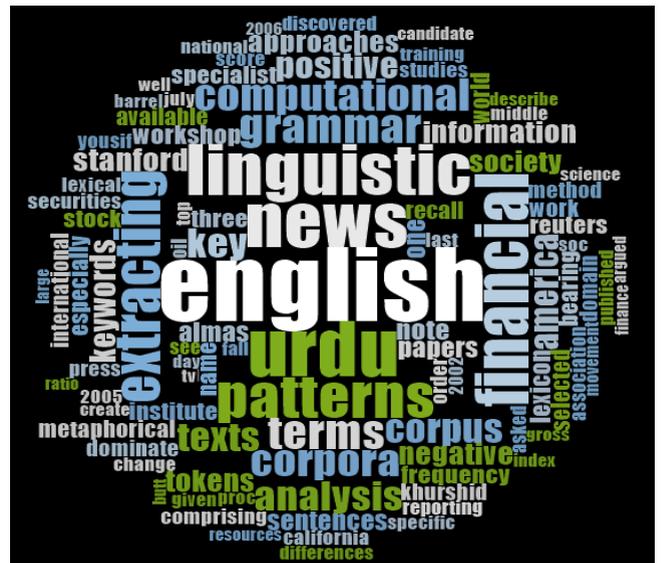

**FIGURE 2b** Word cloud depicting the most frequent words appearing in Step 2 (B) of the ASA screening [244].

### B. OVERVIEW OF INCLUDED STUDIES
The outcomes of our review of selected studies are presented in this section, to highlight the similarities and differences between state-of-the-art ASA studies. Our findings indicate that ASA is applied to a range of various domains, including *health services* [60-62], *telecommunication services* [8, 63, 64], *customers' satisfaction with e-products* [63], *government services* [65], *security* [66], *volunteer work* [67], *politics* [68], [69] *and finance* [70].

Findings also indicate that common approaches used to implement ASA (2002 to 2020) include the *machine learning approach* (Figure 3). It is noted that the *hybrid approach* (mixing *machine learning* with the *unsupervised learning*) first emerged in 2010 and was frequently used until the year 2019. From 2005 to 2020, there was a prevalence of machine learning approaches, specifically *supervised learning*. Between the years 2017 and 2019, the application of machine learning rose drastically, more than other approaches. The use of deep learning has increased rapidly in the research community since 2017. It is evident here that *transfer learning* is still new to the ASA field, with only one study found in 2019 and one study found in 2020.

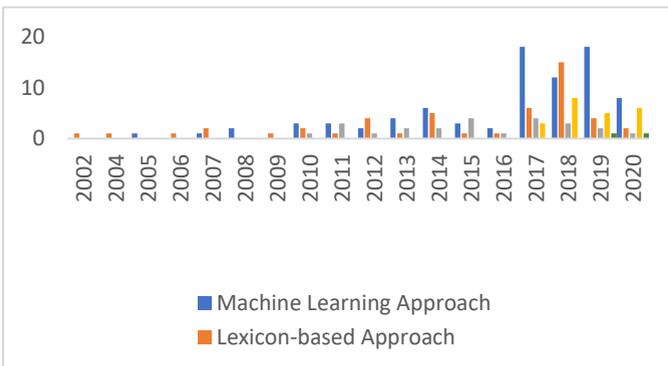

**FIGURE 3:** Distribution of ASA approaches from 2002 to 2020

The following sub-sections provide a detailed analysis of the different approaches identified for ASA.

1) SUPERVISED LEARNING APPROACHES
There are several learning algorithms applied based upon a supervised learning approach [71, 72], which are: *naïve bayes* (NB), *support vector machines* (SVMs), *decision trees* (DTs), *logistic linear regression*, *random forest*, *neural networks* and the *k-nearest neighbours* (K-NN) algorithms. These algorithms were employed as the base classifiers for ASA [26].

A number of experimental studies [26, 50, 62, 73, 74] have used different machine learning algorithms for standard Arabic datasets and dialectal Arabic. For example, [23] evaluated the application of NB and DTs using a multi-dataset in MSA and dialectal Arabic. This dataset consisted of 658 comments from Facebook written in English, 2648 reviews from Aly and Atiya [75] written in MSA, and 409 reviews used by [76]. SA was performed using RapidMiner for a two-way classification (positive, or negative). Evaluation was conducted based on two parameters (accuracy and runtime) with results demonstrating some significance. The two classifiers performed poorly on dialectical Arabic, with 50.76% accuracy for DT and 54.43% for NB. Regarding MSA, the performance was raised to 97.16% with DT and 89.52% with NB. In addition, the performance of both classifiers was enhanced on the English corpus, with 84.25% for NB and 83.87% for DT. They concluded that NB performance is higher on the English corpus, when comparing with dialectical Arabic.

[42] applied different machine learning algorithms NB, AdaBoost, SVM, Ridge Regression (RR), and Maximum Entropy (ME). Their dataset consisted of 151,000 tweets written in the MSA and Egyptian dialect, balanced between positive and negative tweets. Their main finding indicated that RR with Term Frequency -Inverse Document frequency (TF-IDF) as the feature extraction method and 10-fold cross-validation achieved the best result, with 99.90% accuracy, precision, recall and F-measure.

Following this work, [73] compared different algorithms, which are NB, SVM, BNB, Multinomial NB (MNB), Stochastic Gradient Decent (SGD), Logistic Regression (LR), Maximum Entropy (ME), RR, Passive Aggressive (PA), and Adaptive Boosting (Ada-Boost), with different n-gram features using 10-fold cross validation on their dataset Gamal, 2019. Results favoured the unigram feature set, with PA achieving 99.96% for precision, recall, accuracy and F-measure.

In the work of [77], a comparison of linguistic and statistical features was compared between SVM, KNN and ME. Linguistic features included stemming and part-of-speech (POS) tagging, whilst statistical features included TF-IDF. Their dataset consisted of 10,006 tweets labelled with (positive, negative, neutral and objective); where 'objective' means a tweet without opinion and 'neutral' is a tweet with positive and negative opinion in the same tweet. They concluded that SVM outperformed the other classifiers by obtaining 75.21% for precision, 72.15% F-score and 69.33% recall. This implied the suitability of using the suggested features with SVM.

[78] compared SVM and NB applied to 6,921 reviews and comments collected from the Yahoo and Maktoob social networks using 10-fold cross validation and the TF-IDF scheme. They classified comments and reviews into four categories (social, technology, science and arts) then labelled these comments and reviews using three sentiment labels (positive, negative or neutral). Much of the dataset was written in MSA and different Arabic dialects (Egyptian, Khaliji, Levantine, and Arabizi - which is a combination between MSA and English). They concluded that SVM was superior to NB for their unbalanced dataset,

obtaining 64.1% for accuracy and recall and 63.8% for precision.

SVM emerged superior to other algorithms on different datasets, such as that of [79], who proposed the largest offensive words Arabic dataset extracted from Twitter, based on different Arabic dialects. Their dataset contained 10000 tweets labelled with four labels (clean, hate speech, vulgar or offensive). The dataset was evaluated using the following different algorithms: DT, RF, Gaussian NB, Perceptron, AdaBoost, Gradient Boosting, Logistic Regression and SVM, with different pre-trained embedding. They achieved the best F1 at 79.7%, 88.6 recall with SVM using the Mazajak embedding [80].

In addition, [60] proposed a health services dataset written in Arabic, comprising 2026 tweets classified as positive or negative. This area of research applied different Deep and Convolutional Neural Networks and Machine Learning algorithms, such as Logistic Regression, SVM and NB. In addition, they applied the uni-gram and bi-gram as a feature-selection technique and TF-IDF as a weight scheme, with the best accuracy of 91% being achieved by an SVM with Linear Support Vector Classification (LSVC).

[81] assessed the improved chi-square feature selection (ImpCHI), by using SVM and DT on 5070 documents classified into six classes (Business, Entertainment, Middle East, SciTech, Sport and World). SVM with ImpCHI outperformed ImpCHI with DT, obtaining 84.93%, 85.17% and 85.29% for average F-measure, recall and precision. They concluded that when the feature number was between 40 to 900 features, the ImpCHI feature selection outperformed the other feature selections, which were Information gain, Mutual information, and Chi-square.

The same feature algorithm was used by [82], who proposed an approach using a Chi-Square algorithm for feature selection and KNN for classification. They used a Twitter dataset for ASA [83]. It included a perfectly balanced dataset of 2000 tweets, classified as 1000 positive tweets and 1000 negative tweets. They obtained 65.00% using Chi-Square as feature selection and KNN for classification when K=3.

Another study proved the effectiveness of using KNN with ASA [84]. This work proposed an improved K-NN Arabic text classifier using word-level n-grams (unigrams and bigrams) in document indexing and compared this to document indexing based on a single term. They applied their experiment on an Arabic corpus constructed by [85] from online websites and newspapers, with the corpus being placed into the Computer, Economic, Education or Engineering categories. Their approach obtained 87%, 64% and 74% for average precision, recall and F-measure. The study demonstrated that the average accuracy from using n-grams was 74%, while the accuracy from single-term indexing was 67%, thus indicating that the use of n-grams to represent each document provides a higher level of performance, compared to using a single term. In comparison, an alternative study proved that KNN has a poor performance, because of their supposition that a tweet of the same meaning would lead to the same classification [77].

Thus far, we have identified *both SVM and NB as being competitively effective at supervised sentiment classification in the context of Arabic*, [86], [79], and being widely accepted [87]. Some studies have proved the superiority of using NB with Arabic text classifiers, considering it a well-performing algorithm for data mining [88, 89], with a demonstrated accuracy of 82% and 86.5%, respectively, for a macro-averaged precision, and 84.5%, for macro-averaged F-score using the NB classifier. The dataset contained 815 comments written in colloquial Arabic, sourced from two online Saudi newspapers. They manually classified the data, using four labels (strongly positive, positive, negative, and strongly negative). This finding is consistent with that of [90], who concluded that NB provides a higher degree of accuracy than SVM.

On other hand, there exist many studies proving the high accuracy of SVM [87, 91-93]. This is especially true for SA, where SVM was considered not only the best classifier [92, 94] for supervised learning, but also most efficient. [95, 96] demonstrated that, for text classification, the best results were obtained using an SVM, as this does not require parameter tuning. The architecture of SVM - inserting a hyperplane to separate between classified data is explained as being behind its effective performance [95]. In addition, [27] claimed that an SVM was superior to NB regarding accuracy, as there is no reliance upon probabilities and is suitable for high-dimension text. This success has even been reflected in graphical languages, such as Chinese [97, 98]. This presents the possibility of success in other graphical languages, such as Japanese and Arabic. In addition, some studies have stated the reason for the superior SVM performance to be the ability of SVM to handle many classes, and the vectorisation architecture of SVM, which allows for a good quality representation of text [77, 99].

As a result, SVMs have been abundantly applied to movie reviews, while NB has been used within web discourse sites [100]. Furthermore, some studies affirmed that the performance differences between NB and SVM algorithms are based on textual characteristics [101]. Finally, some studies even claimed that SVM is hard to interpret [81].

Concluding, it has been noticed that the *supervised learning approach (particularly machine learning) is the most popular approach for ASA* [102] due to the high-accuracy that it provides using supervised learning [102] and [103]. However, there are some challenges with this approach:
- It requires labelled training data, which is time-consuming and costly [78, 104];
- Due to the need of labelled training data, which requires humans for the annotation, this makes the availability of high-quality datasets slight [104];

- It is domain-dependent because the model performance that was trained on a specific dataset will decrease when trained on a different dataset with a different domain [102, 104];
- It requires a lot of features to differentiate between sentiments [77].

Some studies applied other machine learning techniques (i.e., the unsupervised approach) to identify groups, as is further described in the next section.

### 2) UNSUPERVISED LEARNING APPROACHES
Although the supervised approach has been proven to be superior to the lexicon-based approach [22, 105], it requires for data to be labelled, which is hard to construct. Many studies have attempted to apply instead a lexicon-based approach, with the aim of building an Arabic version [37, 40, 41, 45, 50, 78, 106-111].

The use of lexicon-based approaches differed in the ASA literature. [110] combined the lexicon-based approach and rule-based approach, to propose an Arabic Aspect-based SA. This dataset consisted of 2071 Arabic reviews from government apps. The approach achieved an accuracy and F-measure of 96.57% and 92.50%, respectively. In addition, [111] improved the unsupervised approach for ASA, through the use of valence shifter rules. They applied available lexicons, such as the lexicons proposed by Mohammad et al., [112], ElBeltagy [113], AraSenti-PMI by Al-Twairesh et al., [114], and Arabic Senti-Lexicon by Al-Moslmi et al. [50], etc. The research concluded that the proposed rules enhanced the classification performance by 5%. Moreover, [102] proposed a weighted lexicon-based algorithm (WLBA) of SA for the Saudi dialect. The WLBA concept is to learn from the corpus and not depend upon the lexicon to calculate the weight. The algorithm subtracts the associations between sentiment-bearing and non-sentiment-bearing words, and then, based on the association, it calculates the weights for the words. The researchers applied WLBA to their Saudi dataset for Sentiment Analysis consisting of 4700 tweets. They compared between their proposed approach and two different lexicon-based approaches, the double-polarity approach [115] and the simple algorithm [116]. The simple algorithm method relies on counting, in each sentence, positive and negative words, whilst double polarity depends on the frequency of sentiment words in the sentence. The researchers concluded that WLBA performed better than the double-polarity approach; however, worse than the simple method. They counted some features to enhance the performance, such as supplication (Do'aa), to capture the linguistic complexities of the Arabic language. This provided a performance increased to 85.4% for the average accuracy. Results demonstrated that consideration of linguistic features in ASA is important, and not widely covered within the literature. In addition, the Saudi dataset contained a large amount of Do'aa, therefore supporting its importance of being included within a corpus. Moreover, they proved that there is a strong relation between the sentiment-bearing words and the non-sentiment-bearing words in the Saudi dialect corpus. The same result was obtained by [51] regarding the importance of considering linguistic features, such as negation, for ASA. This research compared a lexicon-based method, a supervised method and a hybrid method. The lexicon-based method relied on counting the positive and negative words. Their proposed approach achieved an accuracy of 91.75%. The same was performed by [107], who compared corpus-based and lexicon-based approaches for ASA. They constructed a lexicon for ASA from a seed of 300 words. Then, they added synonyms to expand the lexicon. After that, they summed up all the weights for the word polarity, including the negation to the weights. They concluded that the lexicon-based method performed inadequately, when the lexicon is small.

As we can see, the lexicon-based approach depends on the creation of a lexicon of good quality [102]. The major advantage of the lexicon-based approach is domain-independence, when constructing a comprehensive lexicon. However, it is hard to construct a comprehensive lexicon [21].

### 2.1 LEXICON BASED APPROACH
For building a lexicon, there are two approaches in the literature [22]: *manual* [117] or *automatic* [118]. The automatic approach includes corpus-based and translation-based approaches [22, 116].

Many researchers applied a manual approach, since the manual approach provided a more accurate lexicon [102, 116]. [37] constructed manually the Sifaat, which is an Arabic lexicon with 3325 adjectives. They subsequently extended it to the Multidialectal Arabic Sentiment Lexicon (SANA). In addition, they proved that the lexicon manually constructed was more accurate than one automatically built; however, researchers are limited in the size of lexicon they may construct [78].

A *dictionary-based approach* depends on using a dictionary to find the synonyms and antonyms of seeds of positive and negative words, until no word is found anymore [51]. One of the drawbacks of using the popular lexicon for translation is this approach is not accurate, due to errors in translation, or cultural variations [119], [120] and [51]. Table 1 in Appendix I shows some sources and dictionaries that were used in previous studies to construct Arabic lexicons. The corpus-based approach depends upon the corpus to generate the polarity words, then uses different approaches, to find the synonyms and antonyms of these words, to generate the lexicon [78]. Some scholars have utilised specific algorithms to construct an automatic lexicon, such as the pointwise mutual information (PMI) statistical method [112]. Following in their steps, [51] offered two sentiment lexicons, AraSenti-Trans and AraSenti-PMI, built from the Twitter dataset Arasenti-tweet [121]. They used two automatic approaches for generating the lexicon and applied a simple lexicon-based approach to evaluate the two lexicons. To generate the first lexicon AraSenTi-Trans, they applied the MADAMIRA tool [122] for pre-processing the dataset. Then, they employed two sentiment lexicons: the Liu lexicon [123] and the MPQA lexicon [19]. The second lexicon was generated using PMI [124], which calculates the association between two terms, in terms of the SA, i.e. the frequency of a word in a positive text, compared to the frequency of the same word in a negative text. AraSenti-Trans includes 131,342 words and AraSenti-PMI includes 93,961 words classified as negative or positive words. They applied a simple lexicon-based approach for evaluating the lexicons on three datasets RR [125], Arasenti-tweet [121], and ASTD [126]. The results showed that the AraSenti-PMI lexicon outperformed the other lexicon. The best F-avg of 88.92% was obtained by the AraSenti-PMI lexicon on the AraSenti-Tweet dataset. Regarding the AraSenti-Trans lexicon, the best F-avg of 59.8% was achieved on ASTD [126]. Compared to the manual building of a lexicon, the automatic approach requires a considerable reduction in effort and ensures that significantly larger lexicons may be produced [78]. Many studies have attempted to apply a lexicon-based approach with the aim of building an Arabic version [37, 40, 41, 45, 50, 78, 97, 107, 108]. A large portion of these studies used manual construction, which provided highly accurate sentiment classification. Some researchers have used available resources, including Arabic WordNet [127], to build MSA lexicons [128, 129]. [130] applied a semi-supervised approach with Arabic WordNet [127] to build a MSA lexicon and achieved 96% classification accuracy. Other researchers have focused on the building of an Arabic dialect lexicon using different approaches (i.e., manual or automatic) to generate lexicons [26, 42, 78, 107, 114, 115, 131]. Although lexicon components have been used successfully for Arabic, a large Saudi dialect lexicon has not yet been fully applied to ASA [102].

2.1.1 Saudi Lexicon

We further critically reviewed the construction of Saudi lexicons. [104] constructed a Saudi dialect sentiment lexicon (SauDiSenti) that consisted of 4431 words and phrases written in MSA and Saudi dialect. It is available online[1]. It is manually constructed from a Saudi dialect twitter corpus (SDTC) [132]. They yielded two annotators to extract the positive and negative terms from the corpus. They extracted 1079 positive terms and 3351 negative terms. For evaluation of the lexicon, they compared it to one of the largest Arabic lexicons, AraSenTi [51]. The result showed that SauDiSenti outperformed AraSenTi, when accounting for the neutral tweets, together with the positive and negative tweets, with 0.437% for the average F-measure. In addition, the AraSenTi outperformed the SauDiSenti, when considering positive and negative tweets, with 0.76 for average F-measure. [102] provided a Saudi dialect lexicon. The lexicon includes 14,000 terms. The lexicon was constructed through three steps: first, the lexicon was expanded using seed words and a learning algorithm [115]. Secondly, the lexicon was built automatically, based on [133]. In the third step, they added new words manually. Moreover, Adayel and Azmi [134] built a Saudi dialect lexicon of 1500 words (500 positive words and 1000 negative words) as a part of the hybrid approach of ASA. They employed SentiWordNet [135] to translate some words to Arabic and assign a sentiment to them. Furthermore, [136] provided a domain-dependent Saudi Stock Market lexicon (SSML). SSML contains 3,861 terms and their sentiment polarities (positive and negative), with two levels of strengths. They constructed the lexicon manually from Twitter and the Saudi shares forum www.saudishares.net/vb/.

TABLE 2
SOURCES USED TO CONSTRUCT ARABIC LEXICONS

| Label | Source for Arabic lexicon | Creator of the source | ASA works using the sources |
|---|---|---|---|
| S1 | SentiWordNet (SWN) | [137] | [37] |
| S2 | General Inquirer (GI) | [138] | [37, 139] |
| S3 | Twitter | N/A | [115] |
| S4 | MPQA Lexicon | [19] | [131] |
| S5 | Liu Lexicon | [123] | [131] |
| S6 | SentiStrength | [140] | [116, 141] |
| S7 | Penn Arabic Treebank | [142] | [93, 106, 139] |
| S8 | Arabic WordNet | [135] | [130] |
| S9 | SWN3 | [143] | [139] |
| S10 | Affect Control Theory Lexicon | [144] | [37] |

2.1.2 Pros and Cons of Lexicon-based Approaches

The lexicon-based approach presents some disadvantages that have been summarised as follows: For aspect-level sentiment analysis, it causes a minimum recall [50]. The lack of training in using the lexicon-based approach is not as effective as the supervised approach for SA [117]. Due to that, the lexicon-based approach depends on the database used; this causes a lack of extensibility[37].

Due to the variety of Arabic dialects, each dialect needs a special lexicon, because of the uniqueness of its lexical information. Therefore, the lexicon-based approach is dialect-dependent, and domain-dependent with the claim that for SA a domain-dependent lexicon outperforms general lexicons [145, 146].

However, there are some advantages from using the lexicon-based approach: No need for model training, which makes it simple [50, 104]. Providing background information via a lexicon with the machine learning training could be optimal [50]. A lexicon-based approach provides the understanding of the impacts of the theoretical framework [147].

3) HYBRID APPROACHES
The hybrid approach is a combination between supervised learning and unsupervised learning [114] [50] and [140].

---
[1] http://corpus.kacst.edu.sa/more_info.jsp

Several studies have found a hybrid approach to be the most suitable technique for SA [50, 106, 114, 141, 148, 149]. Hybrid approaches can contribute to solving the shortcomings of both supervised learning and lexicon-based approaches [150]. The combination of the high accuracy from supervised learning approaches and the flexibility from unsupervised approaches makes hybrid approaches perform the best with SA [50]. Furthermore, some studies proved that the hybrid approach outperforms the supervised approach in accuracy [50].

The hybrid approach is common within the ASA research; for example, [148] applied a lexicon-based approach using SentiWordNet for ASA, alongside with a machine learning classifier. They used SentiWordNet as a feature for SVM. They concluded that using SentiWordNet, as a feature for the SVM algorithm, improved upon the term counting method by raising the accuracy from 65.85% to 69.35%. The same lexicon SentiWordnet was used by [134] to label the tweets; then they used supervised learning, SVM, with n-grams, to classify the text. The results validated the effectiveness of the hybrid approach, with 84% and 84.01% for the F-measure. Additionally, the accuracy was raised, using the hybrid approach over the individual lexicon-based or machine learning approaches.

In another work, [106] proposed SAMAR (a sentence-level ASA for Arabic social media genres). They utilised a polarity lexicon (PL), manually composed of 3982 adjectives, labelled with (positive, negative, or neutral) on the DARDASHA and TAGHREED datasets, to investigate the task of sentence-level construction with MSA and Arabic dialects. TAGREED includes 3015 MSA and dialectal Arabic tweets, while DARDASHA (DAR) includes 2798 Egyptian dialect Arabic chats from the Maktoob website http://chat.mymaktoob.com. They applied PL as a binary feature, to analyse chat or tweets on whether they included positive or negative adjectives. The work concluded that the accuracy of the SA was raised, after applying the polarity lexicon. The best F-score obtained was 95.52%.

Some studies proposed the hybrid approach for ASA, by using an Arabic lexicon to find the sentiment score of the words in a sentence, for example: [114] used rule-based knowledge to be included in a statistical method as a feature. They utilised the AraSenti lexicon [51] as a tweet-score feature for the SVM and NB classifiers. This feature was applied using the AraSenti lexicon. They confirmed the superiority of a hybrid approach with two and three-way classifications. The same hybrid approach was used by [131] for binary ASA, three-way ASA and four-way ASA. The best model performance was 69.9% F-score for binary classification, 61.63% F-score for three-way classification, and 55.07% F-score four-way classification.

Another study used a hybrid approach [151], with a manually built lexicon to define the sentiment scores. They applied the hybrid approach on an Egyptian tweets dataset. They validated their approach on 4800 tweets annotated as positive, negative, or neutral. The results showed that integrating lexical-based features into machine learning enhanced ASA. In addition, [50] proposed an Arabic senti-lexicon, including 3880 terms classified as positive or negative. They utilised the Arabic senti-lexicon to extract the features for machine learning algorithms NB, k-NN, SVMs, logistic linear regression and neural networks. The results demonstrated that feature vectors extracted from an Arabic sentiment lexicon enhanced the classifier performance, with the best macro-F-score of 97.8% favouring logistic linear regression. Similarly, [113] integrated features derived from the NileULex sentiment lexicon [152] onto machine learning algorithms. The datasets used were obtained from social media written in MSA and different Arabic dialectics, such as Saudi, Egyptian, and Levantine. It has been used with the Complement Naïve Bayes (CNB) method [153], which works with unbalanced data. The results showed that using the lexical-based features raised the accuracy of the model.

An interesting concept considered for the hybrid approach was introduced by [141], who presented the hybrid approach as a combined approach, which applied different methods sequentially, to classify the sentiment of a text. It applied two methods to classify Arabic documents, i.e., a lexicon-based one, and a machine learning method, using the maximum entropy followed by a K-NN algorithm on the 8793 Arabic statements found in 1143 posts. The research constructed a lexicon by translating the wordlist from the SentiStrength software [137] from English to Arabic. As Figure 4 shows, the accuracy was raised from 50% to 80% using a combined approach. This hybrid approach was applied later, to examine students' opinion changing in two consecutive semesters.

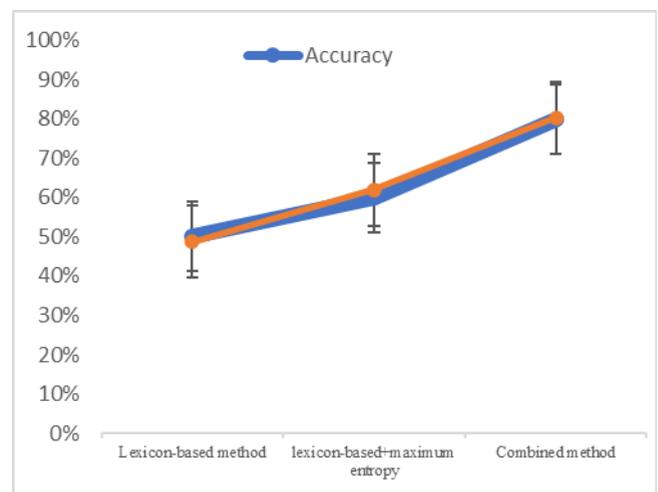

FIGURE 4 Comparison between the three-method performances

[150] proved the outperformance of a hybrid approach over supervised and unsupervised approaches. The research applied a lexicon-based approach to label the dataset of 3000 Saudi dialect tweets. Then, they trained the SVM classifier on the labelled dataset. The hybrid learning results were 96% for

precision, 97% for recall and 90.3% for average accuracy. In addition, [154] used the hybrid approach on the same dataset [150] labelled with the same sentiment lexicon using two machine learning approaches, SVM and K-NN. The results demonstrated the advantage of the hybrid approach over the supervised approach with 90.5% average accuracy using K-NN and 90% average accuracy using SVM.

A similar improvement was demonstrated in another study by [154], who depended on the hybrid approach to switch the sentiment-bearing words with their consistent label in the text. The results showed that the hybrid approach surpassed the corpus-based approach, and the best accuracy (96.34%) was obtained by utilising random forest.

In recent studies, [77] presented the hybrid method as combining linguistic features and statistical features for ASA. POS and stemming were considered as linguistic features, while (TF) and (IDF) were considered as statistical features. They applied SVM, K-NN and ME. The results proved the effectiveness of the hybrid method, additionally, the superiority of SVM over other algorithms, with 72.15% as F-score.

Alternatively, [130] applied semi-supervised learning to evaluate the Arabic lexicon (Arabic SSL). They incorporated Arabic SSL into NB and SVM. They applied their experiment on the OCA corpus [87] and a book review corpus manually collected and annotated. They applied the lexicons to calculate the scores and feed them as features for ASA classifiers. Results demonstrated the superiority of NB over SVM with 97% accuracy. In addition, they concluded that the classification accuracy did not improve using the semi-supervised learning, due to ignoring other factors rather than the sentiment score, such as the order of words within a text.

4) DEEP LEARNING AND TRANSFER LEARNING
To date, deep learning has become a widespread approach in the NLP community [155]. The deep learning mechanism depends upon multiple hidden layers to represent the data, especially with large datasets. Examples of deep learning networks include Convolutional neural networks (CNNs) [156], and Recurrent neural networks (RNNs) [157]. CNN is a feed-forward network mostly applied in computer vision [158]. While RNNs are applied with sequential data.

In the area of sentiment analysis, many scholars proved the deep learning models efficiency [159, 160] and [161]. Recently, a number of studies have investigated the use of deep learning models for ASA [9, 45, 60, 61, 162-177].

CNNs led to good results for many NLP research [62], due to the structural attributes of CNNs. [178] combined CNN with word-embeddings to classify tweets, with results demonstrating the success of this approach. Several researches applied this method, such as [61, 169, 170, 174, 179]. CNNs' capability in choosing excellent features was lauded [61]. In addition, CNN was shown to decrease the number of weights within a model and accordingly decrease complexity [158].

The traditional RNN was seen to struggle during processing of long sequential data [180]. The proposing solution was using the Long Short Memory (LSTM) and Gated Recurrent Unit (GRU) [181, 182] because of the capabilities of the LSTM and GRU in processing long sequential data [61] and in their abilities of inclusiveness in learning – i.e., including the previous output [183].

Thus, the most common RNN models used are Long Short-Term Memory LSTM and Gated Recurrent Unit (GRU) [9, 60, 162-164, 166, 167, 169, 170, 174, 184].

One of the pioneering works using deep learning models for ASA [185] applied a CNN for aspect-based SA for multilingual analysis, as a part of the SemEval-2016 Task 5[186]. Their work obtained an accuracy of 82.72% for ASA. [60] presented a health dataset written in Arabic, which included 2026 tweets classified as positive and negative labels. Different Deep and Convolutional Neural Networks (DNNs and CNNs) and Machine Learning algorithms were used, such as Logistic Regression, SVM and NB. Within this research, the best accuracy was obtained by an SVM with 91%, closely followed by 90% achieved by a CNN. The same dataset was used later by Alayba, et al. [61]. They examined the integration of a CNN and LSTM approach to ASA. The study aimed to improve the ASA accuracy, using their dataset of Arabic health service, with results proving that an integrated approach improved the sentiment classification with 94% for accuracy.

[163] applied a Long Short-Term Memory Recurrent Neural Network (LSTM-RNN) on ASA, using three different Arabic datasets via Twitter: AvaVec, ArabicNews and AraFT. The highest accuracy was achieved by AraFT (93.5%), followed by ArabicNews (91%) and AraVec (88%). The results also demonstrated that a pre-trained word-embedding approach enhanced the performance of the model.

[187] applied gated recurrent unit (GRU-CNN) in ASA using reviews from the web. They employed for text representation the Latent Dirichlet Allocation (LDA) model. The results showed that the GRU-CNN model outperforms the traditional CNN. Their proposed approach obtained 91.64% for the accuracy and F-score. The same method was applied by [188], who employed GRU and CNN on tweets written in the MSA and Arabic dialect. In addition, they used word embeddings. [189] focused on the first task in Semeval-2018, which is about defining the intensity of the sentiment. Their data was obtained from Twitter, in three different languages, including Arabic. They used word-embeddings as a feature. The best results were achieved by CNN, LSTM and Bi-LSTM with avg Pearson r 68.5.

Al-Smadi et al. [167] carried out the application of an LSTM for an aspect-based SA using Arabic reviews of hotels, outperforming the state-of-the-art method. The same steps were followed later by [166], who applied LSTM on aspect-SA with two different settings. The first model is a bidirectional LSTM with a character level and conditional random field (Bi-LSTMCRF). The second one is an aspect-based LSTM. Their dataset contained hotel reviews written in Arabic. They employed two embedding features: character- and word-level. Results demonstrated that their approaches outperformed the state-of-art works.

Recently, pre-trained language models have achieved good results with different NLP target tasks, due to ability of these models to learn with few parameters [190]. Previous approaches depended on features [191]. Open AI GPT [192], a language model built upon the Transformer architecture [193], represents the state-of-art in textual entailment and question-answering [194]. The ULMFiT pre-trained language model [191], composed of three "AWDLSTM" and an LSTM layer [195], is very accurate on different NLP tasks. ULMFiT delivered accurate results for different NLP tasks. The newest language model is BERT [196]. It uses a Transformer network [193]. BERT outperformed the other pre-trained language models, due to its ability to manipulate context from both directions. Another pre-trained language model is RoBERTa [197], which is an enhanced version of the BERT model [196].

The use of transformer language models is still new for ASA studies, with few studies published so far [190, 198, 199]. [198] have used the BERT model [196] on an Arabic tweets dataset. Their generic and sentiment-specific word-embedding model outperformed the BERT model. They explained that this was because the BERT model trained on Wikipedia, which is written in MSA, whereas dialects are used on Twitter. Other research used Arabic word-embedding [200], word2vec model on the AraVec dataset [163, 188] and [189]. While [184] used LSTM and CNN with doc2vec to enhance the performance of SA for a financial site (i.e., StockTwits[2]). The results found that a deep learning approach helped to improve the accuracy of the financial SA.

HULMonA [190] is the first Arabic universal language model; it is based on ULMFiT. It was pretrained on a Large Arabic corpus and fine-tuned to many tasks. It consists of three stages: 1. training AWD-LSTM model [195] on the Arabic Wikipedia corpus, 2. fine-tuning the model on a destination corpus, and 3. for text classification, they included a classification layer in the model. The results showed that hULMonA achieved state-of-art in ASA.

The most recent Arabic universal language model is AraBert [199]. It is a BERT-based model; it was trained on different Arabic datasets. It used the BERT basic configuration [196]. Except, it added a special pre-training prior to the experiment, specific to the Arabic Language. It tried to find the solution for the lexical sparsity in Arabic [179], which is using "ال" "Al" before the word, i.e. a prefix without meaning, by using a Fast and Accurate Arabic Segmenter (Farasa)[201] to segment the words.

5) CORPORA

Compared to other languages, Arabic lacks a large corpus [32, 102, 114, 121, 202]. Many scholars depended on the translation from one language to another to construct their corpus, for example for the Opinion Corpus for Arabic (OCA). It is one of the oldest corpora for ASA by [87], comprising more than 500 Arabic movie reviews. The reviews were translated using automatic machine translation, and the results compared to both Arabic and English versions. Subsequently, most research efforts have focused on enhancing classification accuracy with the OCA dataset [203]. In addition, the MADAR corpus was proposed by [204]. It included 12,000 sentences from Basic Traveling Expression Corpus (BTEC) [205] translated to French, MSA, and 25 Arabic dialects. This corpus for Dialect Identification and Machine Translation is available online[3]. One of the earliest Arabic datasets created as MSA Resource was the Penn Arabic Treebank (PATB) [142]. It consisted of 350,000 words of newswire text. It had 12 parts. This dataset has been a main resource for some state-of-the-art systems and tools, such as MADA [206], and its successor MADAMIRA [122], YAMAMA [207], and [208]. It is available for a fee[4].

Regarding the Arabic dialects, the Egyptian dialect had a lot of attention; one of the earliest Egyptian corpora is the CALLHOME corpus [209]. In addition, Levantine Arabic was researched intensively, leading to the Levantine Arabic Treebank (LATB) [210]. It includes 27,000 words in Jordanian Arabic. Some efforts were made for Tunisian [30] and [211], and Algerian [212]. Regarding the Gulf Arabic corpus, there is the Gumar corpus [213]. It consisted of 1,200 documents written in Gulf Arabic dialects from different forum novels. It is available online[5]. Using the Gumar corpus, a Morphological Corpus of Emirati dialect has been created [214]. This consisted of 200,000 Emirati Arabic dialect words and is freely available[6]. More details about the Arabic corpora are summarised in Appendix II. However, there are shortcomings to the existing corpora and their availability. This is due in part to the strict procedures for gaining permission to reuse aggregated data, with most existing corpora not offering free access. Furthermore, it is clear from Appendix II that the most frequently applied source for Saudi corpora is Twitter. Unfortunately, not all Saudi corpora that were found in the literature are available. In addition, some of them did not mention details about the annotation, which may cause a limitation for using these corpora. Finally, Appendix

---

[2] Stocktwits - The largest community for investors and traders
[3] http://nlp.qatar.cmu.edu/madar/
[4] https://catalog.ldc.upenn.edu/LDC2005T20
[5] https://nyuad.nyu.edu/en/research/centers-labs-and-projects/computational-approaches-to-modeling-language-lab/resources.html
[6] https://nyuad.nyu.edu/en/research/centers-labs-and-projects/computational-approaches-to-modeling-language-lab/resources.html

III illustrates the percentage of different Arabic corpus types. Interestingly, since 2017, we found that dialectal Arabic has been used in more corpora than MSA.

### 6) SYSTEMS AND TOOLS

Many systems and tools that support Arabic are dedicated to Morphological Analysis (MA) [122, 206, 215]. The oldest and pioneering system in this field is the Buckwalter Arabic Morphological Analyzer (BAMA) [215]. It depended on an Arabic dictionary. This dictionary included prefixes, stems and suffixes. [122] proposed MADAMIRA based on two systems: MADA [206] and AMIRA [216]. Many Arabic works were based on BAMA, for example, SAMA 3.1 [217]and MADA + TOKAN [206].

AMIRA [216], another important work, is an Arabic online tool for POS-tagging, tokenisation, and lemmatisation. Another Arabic system is Khoja's Stemmer [218]. It eliminates the prefixes and suffixes from the word, extracting thus the root.

Named Entity Recognition (NER) tools are considered important for extracting semantic features of the text (Benajiba et al., 2008). However, works applying (NER) to Arabic are few [219]. One of the very recent works on NER [76] proposed a real-time named entity recognition system using news from Internet. The F-score for person, location, organisation, noun, and verb was 72.61%, 68.69%, 55.25%, 77.62%, and 65.96%, respectively.

The review of the ASA literature confirmed the effectiveness of techniques (e.g., data mining) for analysing abundant data (i.e., Arabic text) and for projecting patterns for further discussion and analysis (e.g., forecasting). Our review revealed that, although there is an increasing interest in the use of ASA tools, unfortunately, there is no clear recommendation of a reliable enough tool to perform this analysis within a real-world context. In addition, the tools that are widely used in SA field don't support ASA, such as IBM Watson and Hitachi [220], Tableau [221], Micro strategy and Power BI [222]. For this, tech giants like the Saudi Telecom Company are translating Arabic tweets into English and then using Tableau type software to perform SA.

Of the tools that are used for ASA, [223] compared between a created Opinions Polarity Identification (AOPI) tool and two free online SA tools supporting ASA, which are SentiStrength [224] and Social-Mention[7]. They applied them on a corpus including 3,015 opinions written in MSA and different Arabic dialects. The results proved the efficiency of AOPI over the other tools.

The reviewed studies have covered several techniques enabling opinion-oriented information-seeking systems. These highlight the intellectual richness and breadth of the research area. In addition, numerous studies have also proposed many data-mining systems for MSA. Some systems were designed for dialectal Arabic, while others were designed for both MSA and dialects. Appendix IV illustrates and compares the features of these systems within the ASA literature.

The findings of our review indicate that the majority of existing systems for Arabic text used SVM, KNN and NB classifiers, which have proven to be effective with Arabic. However, future research in ASA is expected to adopt other techniques, such as deep learning and neural networks, which has showed already early promise.

## IV. ANALYSIS OF AUTOMATICALLY COLLECTED WIDER LITERATURE ON ASA

Further to our manual, in-depth analysis, we performed an automatic, in-breadth collection of literature in the ASA domain using the CrossRef API[8], Elsevier API[9], Springer API[10], Wiley API[11], Core API[12] and ArXiv preprints repository[13], for an *in-breadth study of the ASA field*.

We applied the search terms used in the manual data collection task discussed earlier to the search facilities provided by the various APIs. The search terms used were: "Arabic sentiment analysis", "Arabic semantic analysis", "Arabic subjective analysis", "Arabic emotion detection", "Arabic text categorization", "Arabic opinion mining", "Arabic lexicon", "Arabic corpora", "Arabic sentiment analysis", "Arabic sentiment classification" and "Arabic Opinion Mining". From these searches we identified a total of 53,405 potentially relevant articles prior to filtering. Given the range of sources used for data collection, there was a large degree of variation in the formatting of the full-text publications.

Filtering was performed on the identified publications at the title and abstract level (as in Step 1 in Section 2.2), to ensure that only relevant documents were used. This was facilitated through conditional Boolean searches of the identified texts, using the search terms used at the API search level. Following filtering, the total number of relevant publications was reduced to 2297. While some platforms provided results in the standard JATS XML[14] format, many returned full-text results in the form of PDF documents which required further conversion into machine readable XML.

Following the collection of the wider literature we followed the framework presented by [225] for the application of the Latent Dirichlet Allocation algorithm to analyse the literature in the domain of ASA. Following the identification of topics within the corpus, we performed a temporal analysis of the changes in topics throughout 2010-2020. Next, we evaluate these changes and compare them to the manual survey review.

---

[7] *Real Time Search - Social Mention*

[8]https://www.crossref.org/education/retrieve-metadata/rest-api/text-and-data-mining-for-researchers/

[9] *https://dev.elsevier.com/*

[10] *https://dev.springernature.com/*

[11]*https://onlinelibrary.wiley.com/library-info/resources/text-and-datamining*

[12] *https://core.ac.uk/services/api/*

[13] *https://pypi.org/project/arxiv/*

[14]*https://jats.nlm.nih.gov/*

### A. TOPIC MODELLING OF WIDER LITERATURE

Given the high number of topics identified by the optimum model, we present ten topics deemed most relevant to the ASA literature study in Appendix V. These demonstrate several distinct areas related to model architecture, including transformer learning (T1), Bayesian learning (T5, T3) and recurrent and convolutional networks (T6). Additionally, the identified distributions place Twitter and Facebook terms into separate topics. There is still, however, a degree of noise present in the topic-word distributions, with some topics containing vague terms (i.e., challenge, multi, support, sad).

Before performing a temporal analysis of the identified word distributions, it was necessary to filter out unrelated or vague terms and, in some cases, provide new terms. Topics which showed a high degree of noise or were not especially relevant were removed at this stage. The final topic-words selected are presented below along with our deductions of their ideal label (see Appendix VI).

### B. TEMPORAL TOPIC ANALYSIS

Following the distillation of suitable terms and filtering of relevant topics we performed a temporal analysis of the automatically identified literature. Figure 6 presents a temporal analysis of model architectures in the ASA domain in the period from 2010 to 2020.

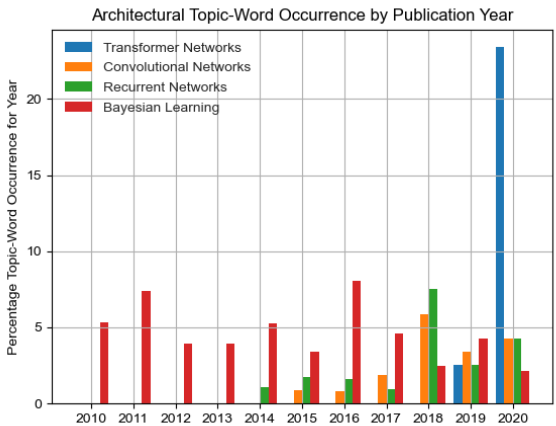

**FIGURE 6:** Proportion of Architecture Topic-Word Occurrences per Year in Automatically Collected Literature

Our temporal topic analysis indicates that Bayesian learning is frequently mentioned in publications throughout our sample, which reflects our identification of several Bayesian classifiers in Appendix IV of our survey. Additionally, the proportion of Bayesian learning in the sample is reduced in years from 2018-2020 in conjunction with the increase in both recurrent and convolutional network occurrences, reflecting our identification of deep learning approaches in Section 3. By acknowledging both convolutional and recurrent network approaches as separate topics we can observe the differences in these. However, there appears no conclusive trend between these in our data. Our analysis of the automatically collected literature identifies transformer networks being present in ASA from 2019, again reinforcing trends identified in Section 3. However, we identified a significant increase in the proportion of literature discussing transformer networks, with over 20% of the corpus mentioning transformer networks in 2020 for our automatically collected data.

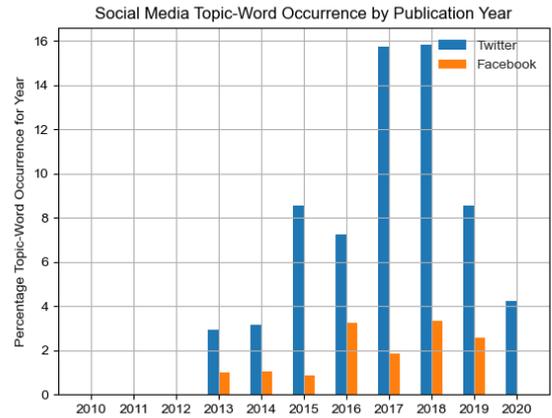

**FIGURE 7:** Proportion of Social Media Topic-Word Occurrences per Year in Automatically Collected Literature

Furthermore, we identify a considerable difference between Facebook and Twitter topic occurrences within the automatically collected literature (Figure 7), with Twitter topic-words being present in more than 15% of publications for the years 2017 and 2018. This can be observed within our study, where a significant proportion of corpora are tweet-based. The prevalence of Twitter within the corpus may be due to the availability of access to real-time streaming data using the Twitter Developer API[15] to allow for collection of tweet data.

---
15 https://developer.twitter.com/en/docs/twitter-api/v1/tweets/filter-realtime/overview

Pre-processing steps taken to ensure fine-grained topic distributions required the removal of frequent terms from the corpus prior to applying LDA [225]. The process of removal of overly frequent terms however, led to the removal of some useful terms, particularly with dialectal terms and abbreviations (e.g., msa, eca). We additionally performed a temporal analysis of the presence of dialectal corpora within the literature, presented in Figure 8.

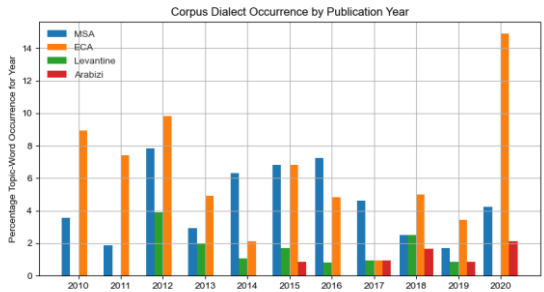

**FIGURE 8:** Proportion of Dialectal Topic-Word Occurrences per Year in Automatically Collected Literature

Analysis indicates that MSA (Modern Standard Arabic) and ECA (Egyptian Colloquial Arabic) are the most common dialects used in corpora. The trends identified in our analysis of the wider literature reflects the prevalence of MSA and Egyptian dialectal corpora in the ASA domain, as found in our study. Additionally, the indication of the Arabizi dialect being present in years 2015-2020 may demonstrate a novel area of ASA research.

## V. DISCUSSION and CONCLUSION

This study reviewed research on ASA to provide a holistic view of the approaches, tools, and resources used in this field. A systematic literature review was conducted, and a total of 133 studies was included in the initial in-depth analysis by a human agent. Further to this, we have reinforced findings from the initial review, through a second analysis of Open Access publications, at a larger scale than manually feasible, through topic modelling and subsequent temporal analysis. The outcome of our review indicated the different approaches used for ASA: machine learning, lexicon-based and hybrid approaches, and corpora-based approaches. In addition, the shortcomings and issues facing ASA research were presented.

Our study offers insight into the issues and challenges associated with ASA research and it provides suggestions for ways to move the field forward. For example, even now, there are many gaps and deficiencies in the studies on ASA. Specifically, Arabic tweets, corpora and datasets for SA are currently only moderately sized. Moreover, Arabic lexicons that have high coverage contain only MSA words and those from Arabic dialects are quite small, with our analysis of wider literature identifying limited applications of Levantine and Arabizi dialects. New corpora need created. Additionally, there is a need to develop ASA tools that can be used in industry as well as academia for Arabic text SA. Hence, we recommend that ASA should be investigated using other techniques, such as deep learning and neural networks in future research.

## Acknowledgements
The authors extend their appreciation to Princess Nourah bint Abdulrahman University. Researchers Supporting Project number (PNURSP2024R349), Princess Nourah bint Abdulrahman University, Riyadh, Saudi Arabia.

# APPENDIXES

## APPENDIX I
### COMPARISON BETWEEN ARABIC LEXICONS [244]

| Lexicon | Lexicon Size | Construction Approach | Source | Reference |
|---|---|---|---|---|
| Arabic senti-lexicon | 3880 terms | A term translation process was revisioned manually | S4 | [50] |
| NileULex | 5953 Egyptian dialectical words and phrases | Manually | [115, 226] | [152] |
| Saudi dialect sentiment lexicon (SauDiSenti) | 4431 words and phrases | Manually | Saudi dialect twitter corpus (SDTC) [132] | [104] |
| large-scale Standard Arabic Sentiment Lexicon (SLSA) | 35,000 lemmas | Machine learning models, with limited using of heuristics | A morphological analyzer for Standard ArabicAraMorph [215] and S1 | [230] |
| Large scale Standard Arabic sentiment lexicon (ArSenL) | 157,969 words | Combination of using Arabic WordNet and an English dictionary | S8, the Standard Arabic Morphological Analyzer (SAMA) [217], English (ESWN) [135] and English WordNet (EWN) [232] | [133] |
| Sifaat | 3,325 adjectives | Manual | S7 | [139] |
| Expanded lexicon | 229,452 entries | Automatic | S1, S9, and Sifaat | [139] |
| Arabic lexicon | 1.8 million phrases | Arabic similarity graph and Manual | Business reviews from web | [108] |
| Polarity lexicon | 3,982 adjectives | Manual | S7 | [37] |
| Egyptian dialect lexicon | 4,392 terms | Manual | S3 | [115] |
| Lexicon that transfers the Jordanian dialect to MSA | 300 words | Manual | Social websites, Internet and chat logs | [96] |
| Lexicon that transfers Arabizi to MSA | N/A | | | |
| Lexicon that transfers emoticons to MSA | N/A | | | |
| Ara-SenTi-Trans | 2.2 million tweets | Automatic | S3 | [51] |
| Ara-SenTi-PMI | | | | |
| Arabic lexicon | 16,800 lexical items | Integration between manual and automatic | S6 | [77] |
| Arabic lexicon | N/A | Manual | S7 | [92] |
| Arabic subjectivity word | 2,600 human-classified comments | Integration between Manual and automatic | S6 and online dictionary | [140] |
| Arabic sentiment lexicon | 7,500 words | Semi-supervised learning | S8 | [130] |
| SANA a dialect Arabic sentiment lexicon | 224,564 entries | Automatic | S1, S2, and S10 | [34] |
| Dialect/slang subjectivity lexicon | 2,000 subjective terms | Automatic | S3 | [233] |
| Idioms/proverbs lexicon for the Egyptian dialect | 32,785 idioms/proverbs | Manual | Arabic websites | [109] |
| Arabic version of SentiStrength | N/A | Automatic | S1 | [225] |

## APPENDIX II
COMPARISON BETWEEN DIFFERENT SAUDI DIALECT CORPORA FOR ASA [244]

| Corpus Name | Ref. | Source | Size | Classification | Online Availability |
|---|---|---|---|---|---|
| AraSenti-Tweet Corpus of Arabic SA | [121] | Twitter | 17,573 tweets | Positive, negative, neutral, or mixed labels. | Not Available |
| Saudi Dialects Twitter Corpus (SDTC) | [132] | Twitter | 5,400 tweets | Positive, negative, neutral, objective, spam, or not sure. | Not Available |
| Sentiment corpus for Saudi dialect | [234] | Twitter | 4000 tweets | Positive or negative. | Not Available |
| Corpus for Sentiment Analysis | [235] | Twitter | 4700 tweets | | Not Available |
| Saudi public opinion | [88] | Two Saudi newspapers | 815 comments | Strongly positive, positive, negative, or strongly negative | Available upon request |
| Saudi corpus | [75] | Twitter | 5,500 tweets | Positive, negative, or neutral | Not Available |
| Saudi corpus | [236] | Twitter | 1,331 tweets | Positive, negative, or neutral | Not Available |

## APPENDIX III
PERCENTAGE OF ARABIC CORPORA OVER TIME BASED ON THE TYPE OF CORPORA [7]

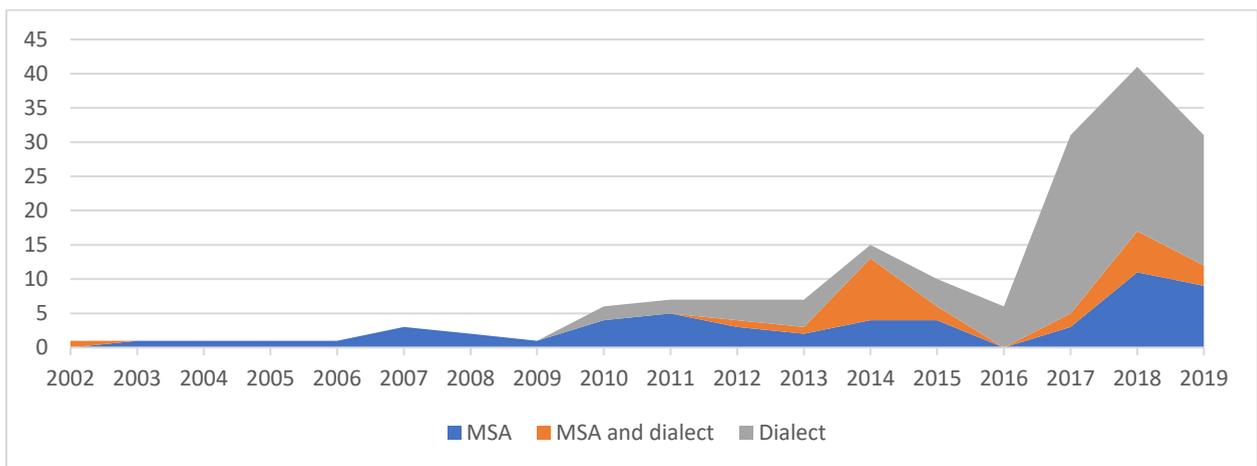

## APPENDIX IV
### DATA-MINING TOOLS USED IN ASA .[244]

| | System | Pre-processing | Algorithms | Data Source | Evaluation |
|---|---|---|---|---|---|
| **ASA tools for Modern Standard Arabic (MSA)** | Standard Arabic sentiment analyzer (SentiArabic) [237] | Yes | lexicon-based combined by a decision tree | SentiTest contained online news and a PATB sentiment annotated by Abdul [40]. | F-score of 76.5% on a blind test set. |
| | [238] | Yes | unsupervised technique | restaurant reviews | 60.5% accuracy |
| | Aara'[88] | Yes | Naïve Bayes classifier | Newspaper comments (Alriyadh and Aljazirah) | F-score was 84.5%. The accuracy of the system is 82% |
| **ASA tools for Dialectal Arabic (DA).** | [239] | Yes | decision tree, support vector machines, and naive bayes | users' comments in Facebook | 73.4% accuracy |
| | [240] | Yes | decision tree, support vector machines, and Naive Bayes | 28300 reviews from YouTube www.youtube.com | 94.5% accuracy |
| | Mazajak[79] | Yes | CNN followed by an LST | SemEval [241] ASTD [171], ArSAS [242] | 92% accuracy |
| | [240] | Yes | decision tree, support vector machines, and Naive Bayes | 28300 reviews from YouTube www.youtube.com | 94.5% accuracy |
| **ASA tools for both MSA and DA Arabic** | Colloquial non-standard Arabic-Modern Standard Arabic Sentiment Analysis (CNSA-MSA-SAT)[243] | Yes | IBK (KNN) Classifier | Arabic reviews and comments from online social website | The accuracy was 90% |
| | SAMAR [106] | Yes | SVM light | Chat websites, social media, web forum and Wikipedia talk pages | The highest accuracy for sentiment classification was for web forum 71.82 |

## APPENDIX V
TOPIC-WORD DISTRIBUTIONS OF MODELS AND OPTIMUM ASSOCIATED PARAMETERS

| Optimal Parameters | Coherence | Topic-Word Distributions of Selected Relevant Topics |
|---|---|---|
| α: 0.01<br>Topics: 201 | 0.39 | T1: arabert, challenge, wordnet, bert, transformer<br>T2: facebook_search, reactions_love, users_trace, sad<br>T3: named_entity, classifier, svm, naive_bayes<br>T4: subjectivity, twitter_feed, supervised, classify<br>T5: gram, multi, mwt_extraction, bayesian, naive_bayes<br>T6: cnn, convolutional, algorithm, lstm, word_embedding<br>T7: code_switching, bayesian, naive_bayes, classifier, complex<br>T8: stemmer, stem, compound, lexicon, text_mine<br>T9: query, sparql, support, rdf, owl<br>T10: finite_state, tree, decision, topic, lsi |

## APPENDIX VI
REFINED TOPIC WORD DISTRIBUTIONS

| Topic No. | Topic-Words | Human Labeled Topic |
|---|---|---|
| 1 | Arabert, bert, transformer, gigabert, transfer_learning | Transformer Networks |
| 2 | CNN, convolutional | Convolutional Networks |
| 3 | Lstm, long short term memory, rnn, recurrent | Recurrent Networks |
| 4 | Bayesian, naive_bayes, bayes | Bayesian Learning |
| 5 | Sparql, rdf, owl | Semantic Web |
| 6 | Twitter, tweet, retweet, twitter_feed | Twitter |
| 7 | Facebook, facebook_post, status_update | Facebook |